# A Novel Face-Anti spoofing Neural Network Model for Face Recognition and Detection

NAME: Soham S. Sarpotdar

College: BITS Pilani, KK Birla Goa Campus

**Abstract**

Face Recognition (FR) systems are being used in a variety of applications, including road crossings, banking, and mobile banking. The widespread use of FR systems has raised concerns about the safety of face biometrics against spoofing attacks, which use the use of a photo or video of a legitimate user's face to gain illegal access to the resources or activities. Despite the development of several FAS or liveness detection methods (which determine whether a face is live or spoofed at the time of acquisition), the problem remains unsolved due to the difficulty of identifying discrimination and operationally reasonably priced spoof characteristics but also approaches. Additionally, certain facial portions are frequently repeated or correlate to image clutter, resulting in poor performance overall. This research proposes a face-anti-spoofing neural network model that outperforms existing models and has an efficiency of 0.89 percent.

**Introduction**

As a result of the rapid spread of Internet technologies, biometrics technology has grown in popularity, and it is now widely utilized for intelligence protection, criminal processes, financial and social stability, therapeutic training, and other fields. Because of its outstanding security, genuineness, and non-contact, the face identification system is more easily accepted by the public than existing biometric identification systems, and it has become an important research avenue for academics and companies[1]. Unauthorized users can utilize face recognition (FR) technology to spread malware, posing a major threat to the system's integrity. As a result, developing a facial anti-spoofing system with improved recognition performance, faster response times, and greater resilience is crucial[2].

Visage anti-spoofing (FAS) detection is a way of evaluating whether a recently obtained facial image is from a living individual or a misleading face. FAS research has received a lot of attention recently, both domestically and internationally, because of its academic importance. The most common spoofing attacks are printing, video replay, and 3D mask attacks. Real and deceptive faces differ in a variety of ways, most notably in image texture data, movement information, and perspective features[3]. Using these characteristics, we may develop numerous FAS systems to distinguish genes between line and counterfeit faces. In recent years, FAS identification research has accelerated, generating a slew of important findings. This research will look at the deep learning (DL) methodology, as well as the technique's benefits and drawbacks, as well as the FAS development trend.

With DL's ongoing progress and outstanding results in the field of FR, an increasing number of investigators have turned to FAS to look into more comprehensive strategies for combating face deception. DL, as opposed to the older manual feature extraction (FE) method, can learn photographs on its own, retrieve more critical and plentiful facial traits, and help differentiate real from phony faces. They begin by proposing a (CNN)[4] to extract features in FAS, paving the path for a new branch of deep learning in the field of FAS[5]. Because the technologies were not yet established, the recognition impact was substantially lower than that of traditional approaches. Furthermore, because of DL's dominance in feature extraction, a lot of research has been done on DL-based FAS. FAS based on DL has improved over time as a result of network improvements, TL[6], due to the unshakable dedication and repeated tries of multiple researchers, has now surpassed the prior technique in terms of a combination of various features and domain generality and has now surpassed the previous technique in terms of domain generality[7].

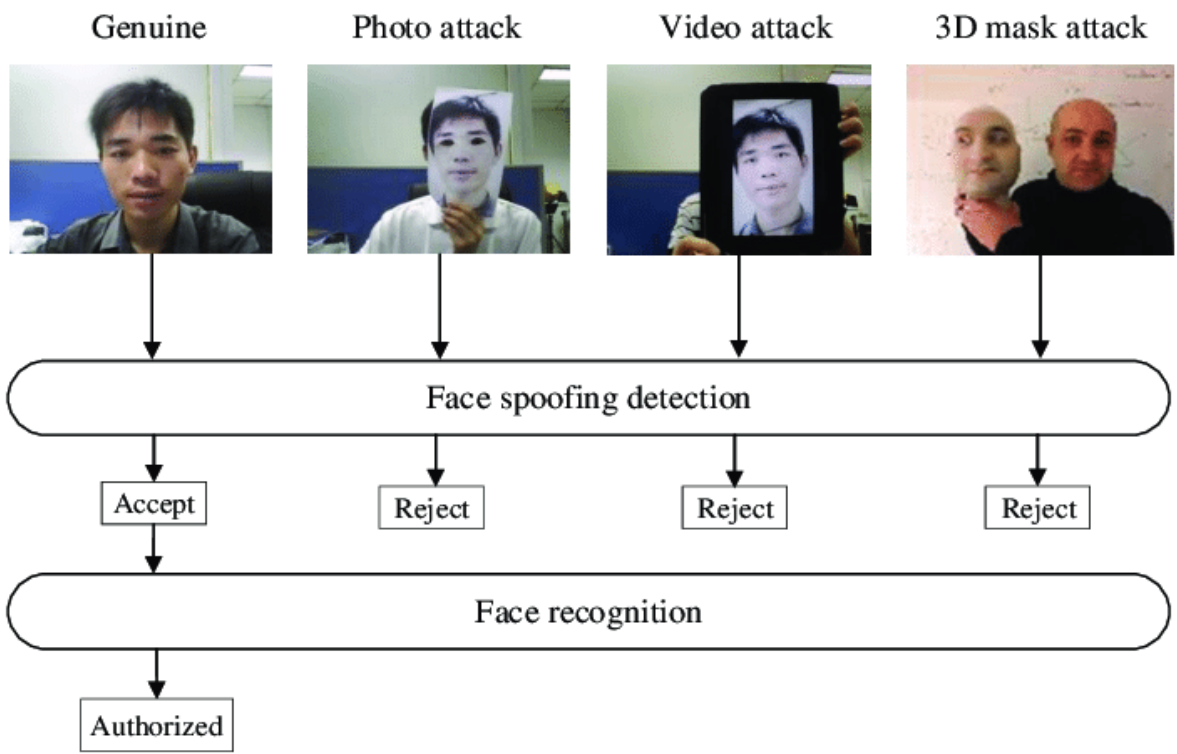

Figure 1. A General framework of face spoofing detection

Face authentication, on the other hand, suffers from the same main flaw as other biometric authentication methods: a nonzero risk of false rejection and acceptance. False rejection is less of an issue because a genuine user can typically try again to be approved, but false acceptance poses a greater security risk. When a fake acceptance happens, the system may be under attack by a malicious impostor trying to get access. Obtaining facial photographs via social media is now easier than ever, allowing attackers to use printed facial photos or recorded facial videos to carry out a variety of attacks. As a result, demand for face presentation detecting attacks (PAD) technology is increasing to maintain the protection of sites that use face recognition systems. Face recognition systems are utilized in places like airports and workplace entrances, as well as in edge device login systems. The face recognition may have access to the server that can do computationally complex computations, or the system may well be equipped with infrared camera equipment. The system, on the other hand, has access to a reduced CPU. As a result, it's only reasonable that the best face PAD algorithm will vary depending on hardware availability[8].

**Related Work**

**Akinori F. Ebihara et al.** had proposed The SpecDiff descriptor is made up of two types of reflections: a specular reflection from the iris region, which has a specific brightness dependent on liveness, and (ii) diffused reflections from the entire face region, which indicate the subject's 3D structure. On an in-house database as well as four publicly available databases, classifiers trained with SpecDiff descriptor beat existing flash-based PAD algorithms: NUAA, Replay-Attack, Impersonating in the Wild, and OULU-NPU. Furthermore, when compared to an end-to-end deep neural network classifier, the proposed technique achieves statistically significantly higher accuracy while having a six-fold faster execution speed. The proposed method's limitations are also assessed under various hostile illumination circumstances to help users implement the system safely. The source code for SpecDiff-spoofing-detector is uploaded on github by Mr. Ebihara and may be found at https://github.com/Akinori-F-Ebihara/SpecDiff-spoofing-detector.
Akinori-F-Ebihara/SpecDiff in-house database sample images are also accessible at https://github.com/Akinori-F-Ebihara/SpecDiff in-house database sample picture[8].

**Wenyun Sun et al.** this study proposed A state-of-the-art face spoofing detection approach that is revised, and it is built on a depth-based Fully Convolutional Network (FCN). Various techniques of supervision are exhaustively studied, including global and local label supervision. For local tasks with inadequate training samples, like the face spoofing detection task, a generic theoretical analysis, and related simulation are offered to demonstrate that local label monitoring is more suitable than global label supervision. The Spatial Aggregation of Bitmap Local Classifiers (SAPLC), which is made of an FCN portion and an aggregation part, is proposed as a result of the investigation. The pixel-level ternary labels, which contain the real foreground, faked foreground, and undecided background, are predicted by the FCN component. The labels are then combined to produce an accurate image-level judgment. Experiments on the CASIA-FASD, Replay-Attack, OULU-NPU and SiW datasets are also carried out to statistically test the proposed SAPLC. The proposed SAPLC outperforms representative deep networks, such as two globally supervised Cnn models, one depth-based FCN, two FCNs with the class label, and two FCNs with tetragonal labels, and achieves superior results close to some state-of-the-art methods under various common protocols, according to the experiments[9].

**S. R. Chavan et al.** study mainly aims to do research into face spoofing techniques and their traceability. This research looks at biometric systems, detecting strategies, and the many characteristics utilized in face spoofing. Face biometric systems are becoming more widely used in almost every field, including government and commercial sectors, as well as on social networking sites such as Facebook, Instagram, WhatsApp, and Google+, where people share personal information such as photos and videos, making them vulnerable to attacks such as face spoofing, DDOS, and phishing. Researchers in the realm of security have been working to build more effective detection methods to protect against spoofing threats. Even though numerous detection algorithms have been created, there are some obstacles in determining the parameters for a face spoofing assault. As a result, greater effort must be put into building more safe and effective detection algorithms for face spoofing assaults[10].

**William Robson et al.** An anti-spoofing system which is based on a set of low-level feature descriptors capable of distinguishing between "live" and "spoof" images and videos has been described. To learn distinguishing traits between the two classes, the suggested technique examines both spatial and temporal data. Experiments using datasets containing photos and videos to validate our technique reveal results that are equivalent to state-of-the-art approaches. Biometrics-based personal identity verification is gaining popularity since it provides for accurate authentication using inherent traits such as the face, voice, iris, fingerprint, and gait. Face recognition techniques, in particular, have been employed in a variety of applications, including security monitoring, access control, crime investigation, and law enforcement. Biometric systems must be resistant to spoofing attempts with images or videos, which are two popular ways to circumvent a face recognition system, to reinforce the findings of verification[11].

**Methodology**

**A. Dataset Gathering**

The proposed method is tested using the CASIA image dataset, which is a widely used and openly available dataset for detecting image forgeries. There are a total of 4795 images, with 1701 genuine and 3274 fakes.

**B. Data Pre-processing**

Pre-processing aims to improve graphic data by masking undesired deformities or enhancing specific graphic features that are useful for further processing and evaluation.

The act of finding and restoring (or deleting) faulty or erroneous data from a record set, table, or database is known as data cleaning. It entails identifying data that is insufficient, incorrect, faulty, or redundant, and then updating, altering, or removing the stale or inaccurate data. Identifying Noisy Images- Image noise is a type of background noise that causes irregular variations in image intensity or color information. It can be done with the image detector and circuitry of a scanner or digital camera. Image noise can be caused by both movie roughness and the inherent impulse noise of an optimum photoelectron. For improved performance, we can check the original and noisy photos in the dataset and convert all of them to error analysis.

Data shuffling attempts to mix up data while preserving logical relationships between columns if required. It randomly rearranges data from a feature's data (for example, a column in pure flat format) or a set of characteristics (e.g. a set of columns). The original and ELA are shown in Figure 2.

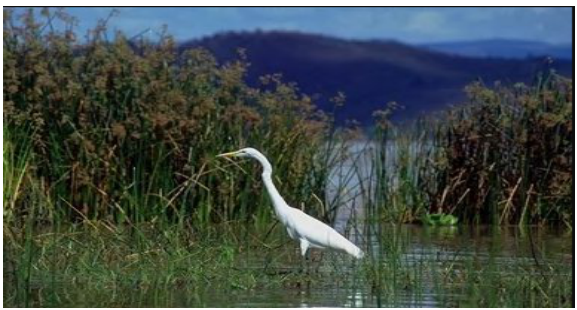 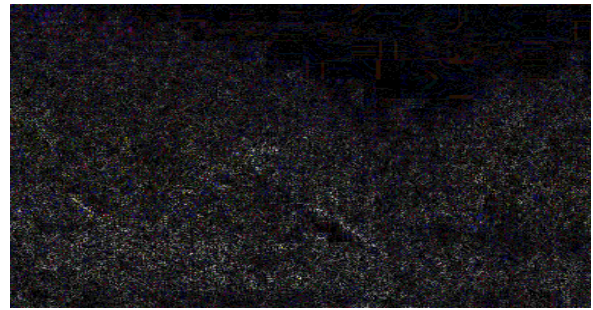

(a) Original Image          (b) ELA Image

Figure 2- Figure showing the original and ELA image.

**C. Model Parameters-**

Conv2D is a 2-D convolution layer that twists a convolution kernel with the layers' data to produce a sequence of outputs[12].

Max-pooling is a type of pooling that selects the largest component from the section of the feature map covered by the filters. As a result, the FM with the most relevant properties of the previous FM would be the output of the max-pooling layer [13].

Dropout Layer- Dropout is a model overfitting prevention method. Dropout consists of setting the outbound edges of hidden nodes (hidden components are made up of neurons) to 0[14] at each iteration of the training step.

Layer- Flattening is the process of transforming data into a 1D array for use in the next layer. To create a single lengthy feature representation, the CL output is flattened. It's also linked to a fully-connected layer, which is the gold standard for classification [15].

A dense layer (DL) in any NN is strongly linked to the layer preceding it, indicating that each of the layer's neurons is related to each other. In ANN, it is the most widely utilized layer.

The DL produces a dimensional array as a result. As a result, the layer is frequently used to modify the vector's dimensionality. These layers also apply transformations to the vector, such as rotation, scaling, and translation[16].

The neural network model is trained sequentially. Figure 2 depicts the used NN model with layers. For false (3331) and real (833) photos, the dataset is divided into 70 percent training and 30 percent testing images. The NN has been trained with the RELU and Sigmoid as activation functions. The network's initial layer is the conv2D layer, which has 2432 parameters, followed by the max-pooling2D layer, and finally the conv2D and max-pooling layers. After then, the dropout, flatten, and dense layers were used in a cascade. The hyperparameters of training using the ADAM optimizer with 20 epochs for a batch size of 32 are shown in Table 1.

Table 1 - Hyperparameters of Training.

| Validation split | 0.2 |
| --- | --- |
| Shuffle | True |
| Epochs | 20 |
| Batch Size | 32 |
| Metrics | Accuracy |
| Loss Function | Binary cross-entropy |
| Optimizer | ADAM |

**D. Performance Metrics-**

**Precision and accuracy**

Accuracy refers to the degree to which a measured value is close to its true value. The degree to which all of the measured values are related is referred to as precision. To put it another way, the proportion of correct categories to total classifications is the measure of accuracy.

**Recall/Sensitivity**

The proportion of true positives to the total number of actual positives is known as sensitivity. Similarly, specificity is the proportion of genuine negatives to total negatives, often known as the true negative rate[17].

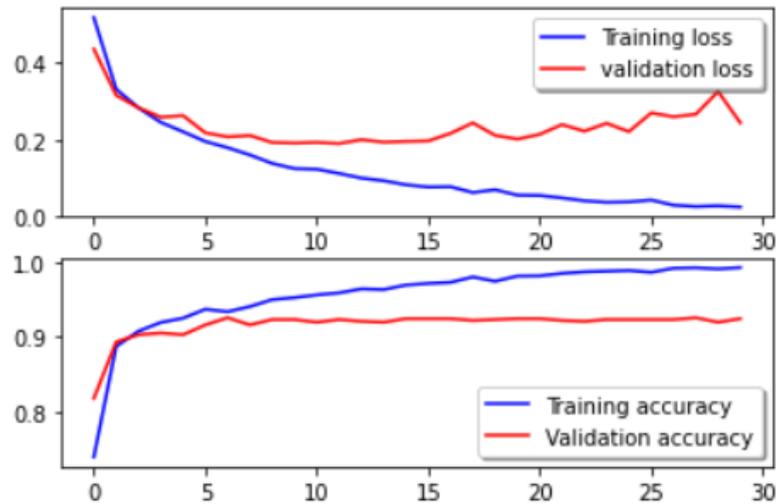

Figure 3- Figure showing the loss and accuracy graph.

**F1-Score**

We offer the F1 score as a statistic that provides a better indication of examples that have been incorrectly classified. When a model's accuracy is greater than 90%, it is considered accurate. This is calculated using the harmonic mean of precision and recall. Accuracy is used when TP and TN are more important. The F1 score is a superior statistic when the class distribution is unequal and FP and FN are more important[18]. The formulas for all of the metrics are listed below.

Table 2 – Table showing a comparison of the base and the proposed results.

| Results | Recall | Precision | F1 Score | Accuracy |
|---|---|---|---|---|
| **SVM** | 0.81 | 0.86 | 0.81 | 0.85 |
| **Propose DNN** | 0.83 | 0.97 | 0.90 | 0.89 |

The accuracy and loss graphs for the evaluated findings are shown in Figure 3. The proposed system accuracy of 0.89 is shown in Table 2 as a comparison of the base and proposed results. The proposed model's accuracy, recall, and f1-score are 0.97, 0.83, and 0.90, respectively. The generated photos' false and actual confidence are presented below.

**Predictions-**

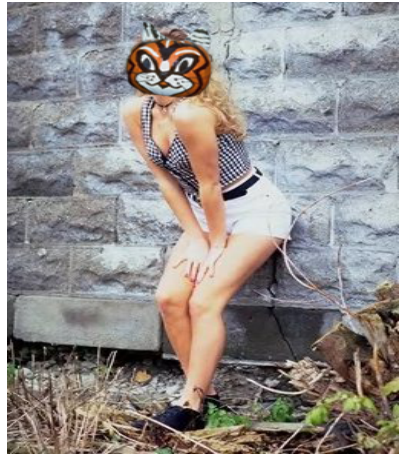 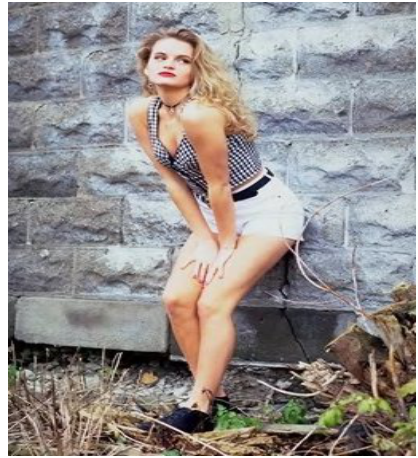

Class: Fake Confidence: 96.67   Class: Real Confidence: 99.80

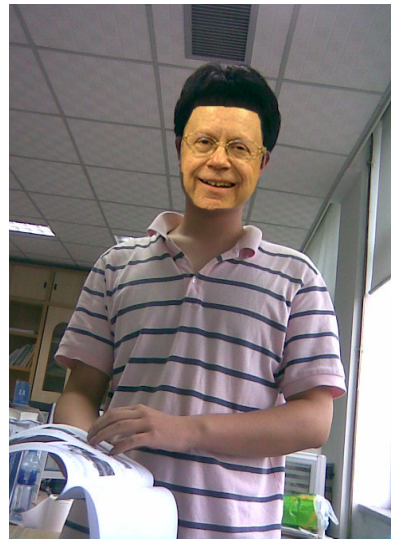 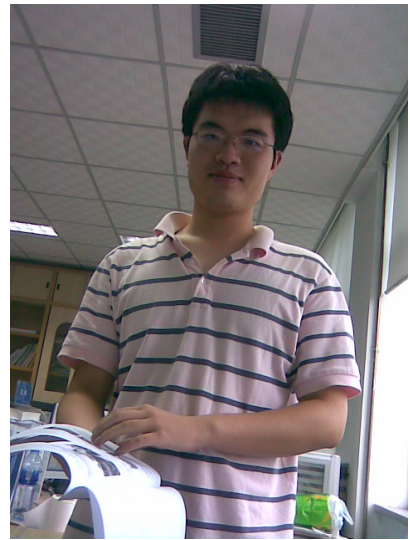

Class: Fake Confidence: 98.75   Class: Fake Confidence: 98.11

**Conclusion**

As AI has grown more frequently employed in real life, FR has become an important tool for achieving protection. In the fight against destructive attacks, FAS has become a major issue. From the beginning of manual FE methodologies based on the image texture, picture quality, and extract features, to using DL to instantly features extracted, merged with connectivity up-gradation, feature integration, and domain generalization, the research of face impersonating identification has indeed been monitored continuously and revised, and the effectiveness and efficiency of an identifier have now reached a significant state.

**References**

[1]  S. Hashemifard and M. Akbari, "A Compact Deep Learning Model for Face Spoofing Detection," 2021.

[2]  N. Al-Huda Taha, T. M. Hassan, and M. A. Younis, "Face Spoofing Detection Using Deep CNN," *Turkish J. Comput. Math. Educ.*, vol. 12, no. 13, pp. 4363–4373, 2021.

[3]  Z. Ming, M. Visani, M. M. Luqman, and J. C. Burie, "A survey on anti-spoofing methods for facial recognition with RGB cameras of generic consumer devices," *J. Imaging*, vol. 6, no. 12, 2020, doi: 10.3390/jimaging6120139.

[4]  Y. Liang, C. Hong, and W. Zhuang, "Face spoof attack detection with hypergraph capsule convolutional neural networks," *Int. J. Comput. Intell. Syst.*, vol. 14, no. 1, pp. 1396–1402, 2021, doi: 10.2991/IJCIS.D.210419.003.

[5]  G. B. De Souza, J. P. Papa, and A. N. Marana, "Efficient Deep Learning Architectures for Face Presentation Attack Detection," pp. 112–118, 2021, doi: 10.5753/sibgrapi.est.2020.12992.

[6]  I. A. Khalid, "Transfer Learning for Image Classification using Tensorflow," *towards data science*, 2020. .

[7]  M. Zhang, K. Zeng, and J. Wang, "A Survey on Face Anti-Spoofing Algorithms," *J. Inf. Hiding Priv. Prot.*, vol. 2, no. 1, pp. 21–34, 2020, doi: 10.32604/jihpp.2020.010467.

[8]  A. F. Ebihara, K. Sakurai, and H. Imaoka, "Efficient Face Spoofing Detection with Flash," *IEEE Trans. Biometrics, Behav. Identity Sci.*, vol. 3, no. 4, pp. 535–549, 2021, doi: 10.1109/TBIOM.2021.3076816.

[9]  W. Sun, Y. Song, C. Chen, J. Huang, and A. C. Kot, "Face Spoofing Detection Based on Local Ternary Label Supervision in Fully Convolutional Networks," *IEEE Trans. Inf. Forensics Secur.*, vol. 15, pp. 3181–3196, 2020, doi: 10.1109/TIFS.2020.2985530.

[10] S. R. Chavan, S. S. Sherekar, and V. M. Thakre, "Traceability Analysis of Face Spoofing Detection Techniques Using Machine Learning," *Proceeding - 1st Int. Conf. Innov. Trends Adv. Eng. Technol. ICITAET 2019*, pp. 84–88, 2019, doi: 10.1109/ICITAET47105.2019.9170212.

[11] W. R. Schwartz, A. Rocha, and H. Pedrini, "Face Spoofing Detection through Partial Least Squares," *Int. Jt. Conf. Biometrics*, 2011.

[12] S. Pouyanfar *et al.*, "A Survey on Deep Learning," *ACM Comput. Surv.*, vol. 51, no. 5, pp. 1–36, 2019, doi: 10.1145/3234150.

[13] K. L. Masita, A. N. Hasan, and T. Shongwe, "Deep learning in object detection: A review," *2020 Int. Conf. Artif. Intell. Big Data, Comput. Data Commun. Syst. icABCD 2020 - Proc.*, no. August, 2020, doi: 10.1109/icABCD49160.2020.9183866.

[14] B. U. Manalu, Tulus, and S. Efendi, "Deep learning performance in sentiment analysis," *2020 4th Int. Conf. Electr. Telecommun. Comput. Eng. ELTICOM 2020 - Proc.*, pp. 97–102, 2020, doi: 10.1109/ELTICOM50775.2020.9230488.

[15] A. P. Shirahatti, "A Survey of Deep Learning for Sentiment Analysis," vol. V, no. I, pp. 1–7.

[16] W. Weng and X. Zhu, "INet: Convolutional Networks for Biomedical Image Segmentation," *IEEE Access*, vol. 9, pp. 16591–16603, 2021, doi: 10.1109/ACCESS.2021.3053408.

[17] ML, "Classification: Precision and Recall," *machine learning crash course*, 2020. .

[18] R, "ROC Curves," *machine learning crash course*, 2020. .